\documentclass[journal]{IEEEtran}
\pdfoutput=1
\usepackage{cite}
\ifCLASSINFOpdf
\else
\fi
\usepackage{stix}
\newcommand{\field}[1]{\mathbb{#1}}

\newcommand{\R}{\field{R}} 
\newcommand{\commentOut}[1]{}
\newcommand{\Th}{^{\text{th}}}

\usepackage{mathrsfs}
\usepackage{subcaption}
\usepackage{graphicx}
\usepackage{svg}
\usepackage{amsmath,amssymb,amsthm}
\usepackage{algorithm}
\usepackage{comment}
\usepackage{float}
\usepackage{array}
\usepackage{algpseudocode}
\usepackage{makecell}
\usepackage{graphicx}
\usepackage{multirow}
\usepackage{balance}
\usepackage{dirtytalk}

\makeatletter
\newcommand*\titleheader[1]{\gdef\@titleheader{#1}}
\AtBeginDocument{%
\let\st@red@title\@title
\def\@title{%
\bgroup\normalfont\large\centering\@titleheader\par\egroup
\vskip1.5em\st@red@title}
}
\makeatother

\makeatletter
\def\endthebibliography{%
	\def\@noitemerr{\@latex@warning{Empty `thebibliography' environment}}%
	\endlist
}
\makeatother

\begin{document}
	\title{A Semi-Supervised Approach for  Power System Event Identification}
	
	\author{Nima~Taghipourbazargani,~\IEEEmembership{Member,~IEEE,}
		Lalitha~Sankar,~\IEEEmembership{Senior~Member,~IEEE,}
		and~Oliver~Kosut,~\IEEEmembership{Senior~Member,~IEEE}
		\thanks{This work was supported in parts by the National Science Foundation under Grants OAC-1934766, EPCN-2246658, the Power System Engineering Research Center (PSERC) under Project S-87, and the U.S.-Israel Energy Center managed by the Israel-U.S. Binational Industrial Research and Development (BIRD) Foundation. \\ The authors are with the School of Electrical, Computer, and Energy
			Engineering, Arizona State University, Tempe, AZ 85281 USA (e-mail:
			\{ntaghip1,lalithasankar,okosut\}@asu.edu).}}

	
	\maketitle

\begin{abstract}\label{sec: abs}
Event identification is increasingly crucial for enhancing the reliability, security, and stability of the electric power system. With the growing deployment of Phasor Measurement Units (PMUs) and advancements in data science, there are promising opportunities to explore data-driven event identification via machine learning classification techniques. However, obtaining accurately labeled eventful PMU data samples remains challenging due to its labor-intensive nature and real-time uncertainty about the event type. Thus, it is natural to use semi-supervised learning techniques, which make use of both labeled and unlabeled samples. 
We evaluate three categories of classical semi-supervised approaches: (i) self-training, (ii) transductive support vector machines (TSVM), and (iii) graph-based label spreading (LS) method. Our approach characterizes events using physically interpretable features extracted from modal analysis of synthetic eventful PMU data. We focus on identifying four event classes (load loss, generation loss, line trip, and bus fault) critical for grid operations. We developed and publicly shared a comprehensive Event Identification package comprising data generation, feature extraction, and event identification with limited labels using semi-supervised methodologies. Using this package, we generate and evaluate eventful PMU data for the South Carolina 500-Bus synthetic network. Our evaluation consistently shows that graph-based LS outperforms the other methods, significantly improving event identification performance with a small number of labeled samples.

\end{abstract}

	\begin{IEEEkeywords}
		Event identification, semi-supervised learning, phasor measurement units, mode decomposition.  
	\end{IEEEkeywords}
\IEEEpeerreviewmaketitle

\section{Introduction}\label{sec: I_Intro}

\IEEEPARstart{P}{ower} systems are inherently complex dynamical systems, primarily due to the involvement of diverse components such as generators, buses, lines, and loads with varying sizes, all exhibiting non-linear behavior and intricate interactions. Given their extensive geographical coverage and scale, power systems are susceptible to various classes of events (for example, generation loss, load loss, line trips). Event identification methods can be used in real-time operations to guide control actions, as well as for off-line analysis of past events to make the system more reliable, secure, and stable in the future.
Numerous studies have explored event detection in power systems, focusing on determining whether an event has occurred \cite{Detection, Unsupervised_YangWeng, Detection2}. However, event identification, which involves discerning the specific class of event that occurred, presents even greater challenges since it requires learning the unique signatures of different events.
Our primary focus here is on addressing the event identification problem in power systems. To this end, our analysis in the sequel assumes that an event has been detected with knowledge of its precise time.

\textit{Related Work}: The increasing deployment of Phasor Measurement Units (PMUs) throughout the grid, coupled with advancements in machine learning techniques, presents invaluable opportunities for exploring advanced data-driven event identification methods. These methods offer the distinct advantage of differentiating between various classes of power system events based on high-dimensional spatio-temporally correlated time-synchronized phasor measurements with high resolution, without heavily relying on dynamic modeling of power system components. 
The majority of the recent literature in the field of data-driven event identification (e.g., \cite{SDT-Vittal, Supervised6-CNN,Supervised2-KNN,Supervised3-SVM,Supervised4-ELM,CNN_new1,TextMining_Sara,DNN_new,nn_2_lstm_new}) employs machine learning and pattern recognition techniques for making statistical inferences or decisions using system measurements. Most of these studies use supervised learning, which requires accurate labeled data.
However, acquiring expert knowledge for labeling various classes of events can be expensive and laborious \cite{EI_2}. 
Such limitations motivate researchers to leverage simulation-based synthetic eventful PMU data for investigating and evaluating the performance of their proposed event identification methods (e.g., \cite{kernels_NN_TSG, Adverial_SS_TSG, Supervised4-ELM}). 
Despite the availability of several resources providing access to synthetic test cases with transmission and/or distribution network models for dynamic simulations \cite{TX500_1, TX500_2}, conducting a fair comparison of diverse event identification methods pose a significant challenge. This challenge primarily stems from the numerous  parameters associated with dynamic models of system components and simulation settings, coupled with the diverse characteristics of simulated events, such as class, duration, and location.
While certain recent publications may have access to significant real and/or synthetic PMU data (e.g., \cite{sslei_2, Partial-Arghandeh}), the lack of publicly available properly labeled eventful PMU data continues to be a persistent concern.

Unsupervised and semi-supervised learning are common practices in machine learning when dealing with no or limited labeled data. Unsupervised learning aims to infer the underlying structure based only on unlabeled data. Although unsupervised learning algorithms can distinguish between clusters of events \cite{Unsupervised_YangWeng,Unsupervised1-Ellipsoid,Unsupervised3-Ensemble,Unsupervised4-Kmeans,Unsupervised2-PCA, Yangweng_limitedlabel}, they do not possess the ground truth to associate each cluster with its real-world meaning.
Semi-supervised learning approaches, on the other hand, aim to label unlabeled data points using knowledge learned from a small number of labeled data points, which can significantly enhance the performance of a classification task \cite{SS-approaches}.
A a framework for event detection, localization, and classification in power grids based on semi-supervised learning is presented in \cite{sslei_2}. A pseudo-labeling (PL) technique is adopted to classify events using the convolutional neural network (CNN) backbone with cross-entropy loss.
A semi-supervised event identification framework is proposed in \cite{Haoran_limited_label_SGSMA} which utilizes a hybrid machine learning-based method to reduce biases of different classifiers.
In \cite{sslei_1}, the authors explore the application of deep learning techniques and PMU data to develop real-time event identification models for transmission networks. 
This is achieved by leveraging information from a large pool of unlabeled events, while also taking into account the class distribution mismatch problem.
In \cite{Arghandeh_partial_TSG}, the authors proposed hidden structure semi-supervised machine (HS3M), a novel data-driven event identification method that combines unlabeled and partially labeled data to address limitations in supervised, semi-supervised, and hidden structure learning. 
 The existing literature on neural network-based event identification methods is marked by certain limitations and challenges. These encompass
restricted interpretability in feature extraction, elevated computational intractability, and the necessity for meticulous parameter calibration.
Moreover, it is worth noting that, to the best of the authors' knowledge, a thorough investigation into the effects of the initial distribution of labeled and unlabeled samples has not been undertaken.
This paper introduces a semi-supervised event identification framework to explore the potential benefits of incorporating unlabeled samples in enhancing the performance of the event identification task. 
To this end, we thoroughly investigate and compare the performance of various semi-supervised algorithms, including: (i) self-training with different base classifiers (i.e.,  support vector machine with linear kernel (SVML) as well as with radial basis function kernel (SVMR), gradient boosting (GB), decision trees (DT), and $K$-Nearest Neighbors ($K$NN)), (ii) transductive support vector machines (TSVM), and (iii) graph-based label spreading (LS) to explore their effectiveness. 
We chose these classical semi-supervised models for two primary reasons: firstly, the wide array of proposed semi-supervised classification algorithms in the past two decades (see, \cite{surveySSL}, and references therein) necessitates a comprehensive understanding of which models are most suitable and efficient for event identification; and secondly, they provide a more clear illustration and intuition of the impact of incorporating unlabeled samples compared to more advanced methods. Although there may not be a one-size-fits-all solution, each method has its own advantages and disadvantages, and it is important to evaluate their suitability.
Notably, our experiments consistently illustrate the superior performance of the graph-based LS method compared to other approaches. Even in worst-case scenarios where the initial distribution of labeled and unlabeled samples does not necessarily reflect the true distribution of event classes, the graph-based LS method stands out in robustly and significantly enhancing event identification performance.
Our key contributions are as follows: 
\begin{itemize}
    \item Introduction of a semi-supervised event identification framework that leverages physically interpretable features derived from modal analysis of PMU data.
    \item Thorough exploration of the influence of the initial distribution of labeled and unlabeled samples, along with the quantity of unlabeled samples, on the efficacy of diverse semi-supervised event identification techniques.
    \item Development of an all-inclusive Event Identification package\footnote{https://github.com/SankarLab/PSMLEI-public} comprising of an event generation module based on the power system simulator for engineering (PSS$^{\circledR}$E) Python application programming interface (API), a feature extraction module utilizing methodologies from our previous research \cite{NT_TPS}, and a semi-supervised classification module.
\end{itemize}




The remainder of the paper is organized as follows. Section~\ref{sec: II_data_generation} describes the simulation process to generate the synthetic eventful PMU data. We explain the proposed semi-supervised event identification framework in Section~\ref{sec: SS_Framework}. In Section~\ref{sec: Models}, we further elaborate on the pseudo-labeling process of the unlabeled samples, and the classification models. We discuss the simulation results in Section~\ref{sec: sim_result}. Finally, Section~\ref{sec: s6_Conclusion} concludes the paper.

\ifCLASSOPTIONcaptionsoff
\newpage
\fi

\section{Generation of the Synthetic Eventful Time-series PMU Data}\label{sec: II_data_generation}

Consider an electric grid composed of set of loads, generators, lines, and buses. 
We investigate four distinct event classes denoted as $\mathcal{E} \in \{\text{LL, GL, LT, BF}\}$, representing load loss, generation loss, line trip, and bus fault events, respectively. 
Each PMU provides multiple measurement channels relative to its installation bus. In this study, we focus on voltage magnitude ($V_m$), corresponding angle ($V_a$), and frequency ($F$) channels for clarity, with potential inclusion of other channels. For any channel $c \in \mathcal{C} = \{V_m, V_a, F\}$, let $y^c_i(n)\in\R$ represent the $n^{\text{th}}$ measurement, $n=0,\ldots,N-1$, where the total number of samples is $N$, from the $i\Th$ PMU. Assuming PMU sampling period of $T_s$, we thus collect eventful data for $t_s=NT_s$ seconds. 
These measurements, for the $c\Th$ channel, are collated from $m$ PMUs to form a matrix $\mathbf{\mathcal{Y}^c}= [\cdots, \mathbf{y}^{c}_i ,\cdots]^T \in \R^{m \times N}$ where $\mathbf{y}^{c}_i$ is a $N$-length (column) vector for the $i\Th$ PMU with entries $y^c_i(n)$, for all $n$. We use superscript $T$ to denote the tranpose operator. 
Finally, for each event, we define $\mathbf{\mathcal{M}} = [[\mathcal{Y}^{V_m}]^T, [\mathcal{Y}^{V_a}]^T, [\mathcal{Y}^{F}]^T ]^T \in\R^{|\mathcal{C}| m \times N}$ by aggregating all the phasor measurements from $m$ PMUs, $3$ channels, and for $N$ samples.

Within this setting, we develop a publicly available Python code which leverages the PSS$^{\circledR}$E software Python Application Program Interface (API) to generate synthetic eventful PMU data.
To ensure a realistic and diverse dataset, we consider the following 
two steps: 
Firstly, we linearly adjust all loads  by a factor that ranges from 95\% to 105\% of their normal loading conditions. Secondly, we add zero-mean random fluctuations, ranging from $\pm 2\%$ of the adjusted loads, to simulate unpredictable variations observed in real-world power systems.\footnote{The load change intervals specified in this paper can be adjusted depending on the stability of the system under study, ensuring that the system can return to an acceptable state of equilibrium following a disturbance.} 
To generate eventful data, for each system component and loading condition considered, 
we employ the following systematic approach:
(i) We begin by applying a new initial loading condition to each load in the system; a power flow analysis for this setting then gives us the initial state conditions for the next step. 
(ii) We use this initial condition to initiate a $t_f$-second flat run dynamic simulatsion.
(iii) At the $t_f$ second, we introduce a disturbance (i.e., LL, GL, and LT)  to a selected component. For BF events, we clear the disturbance after $t_{\text{clr}}$ seconds. 
(iv) Finally, we model the event simulation for additional $t_s$ seconds which then allows us 
create the data matrix $\mathbf{\mathcal{M}}$, representing the PMU measurements associated with the simulated event.
We repeat this procedure to generate a desired number of events for each event type.

\subsection{Generating Event Features Using Modal Analysis}\label{sec: II_B_event_data}

The first step in identifying a system event is to extract a set of delineating features that are likely to contain information regarding the event class. Using the fact that temporal effects in a power system are driven by the interacting dynamics of system components, we use mode decomposition to extract features. More specifically, we assume that each PMU data stream after an event consists of a superposition of a small number of dominant dynamic modes. The resulting features include the frequency and damping ratio of each mode, as well as the residual coefficients indicating the quantity of each mode present. We briefly summarize the mathematical model and refer readers to our recent work \cite{NT_TPS} for additional details.

We assume that $y^c_i(n)$ after an event consists of a superposition of $p$ common damped sinusoidal modes as 
\begin{equation}\label{eq:modalrep}
y^c_i(n) = \sum_{k=1}^{p}  R^c_{k,i} \times (Z^c_k)^n  + \epsilon^c_i(n), \quad i  \in \{1, \ldots , m\},  \quad c \in \mathcal{C}
\end{equation}
where for any given channel $c \in \mathcal{C}$,  $\epsilon^c_i(n)$ represents the noise in the $i\Th$ PMU measurement and $Z^c_k$ is the $k\Th$ mode associated with the event. We represent each mode as $Z^c_k = \exp(\lambda^c_k T_s)$ where $\lambda^c_k= \sigma^c_k \pm j \omega^c_k$ and $\sigma^c_k$ and $\omega^c_k$ are the damping factor and angular frequency of the $k\Th$ mode, respectively. The residue $R^c_{k,i}$  of the $k\Th$ mode for the $i\Th$ PMU  is defined by its magnitude $|R^c_{k,i}|$ and angle $\theta^c_{k,i}$. 
For any given channel $c$, typically a small subset of the PMUs ($m'< m$) capture the dynamic response of the system after an event. 
Thus, we only keep the residues of a set of $m'$ PMUs with the largest magnitudes. 
Note that the $m'$ PMUs are not necessarily the same PMUs for different events (see \cite{NT_TPS} for further details).

Using the above procedure, for each channel $c$, we define a row vector of features, $\mathcal{F}^{c}$, of length $2p(m'  + 1)$ as:   
\begin{equation}\label{eq: VPM}
 \mathcal{F}^{c} =  \big[\{\omega^c_k\}_{k=1}^{p},
 \{\sigma^c_k\}_{k=1}^{p},
    \{|R^c_{k,i}|\}_{k=1}^{p},
    \{\theta^c_{k,i}\}_{k=1}^{p} \big]_{i \in \{1, \cdots , m'\}} 
 \end{equation}
which consists of $p$ angular frequencies, $p$ damping factors and the corresponding magnitude and angle of the residues for each of the $m'$ PMUs (with the largest residue magnitudes) and the $p$ modes. 

\subsection{Generating the overall dataset}\label{sec: II_B_vec_struct}

Let $n_D$ be the total number of events generated over all event classes. 
Following modal analysis on the PMU measurements as described above, we can represent the $i\Th$ event, $i \in \mathcal{I}_D = \{1,\ldots,n_D\}$, as a $d=2p|\mathcal{C}|(m'  + 1)$-length vector  $x_i^T = [\mathcal{F}^{V_m},\mathcal{F}^{V_a}, \mathcal{F}^{F}]$.
Considering a positive integer $j \in \{1, \cdots, |\mathcal{E}|\}$ as an event label, we associate a one-hot encoded vector $y_i \in \R^{|\mathcal{E}|}$, where  $|\mathcal{E}|$ is the total number of event classes, $y_{ij} = 1$ if $x_i$ is labeled as event $j$, and $y_{ij} = 0$, for the unlabeled event $x_i$.


Collating the events and labels from all event classes, we obtain a data matrix $\mathbf{D} = \{\mathbf{X}_D, \mathbf{Y}_D\}$ where 
$\mathbf{X}_D = [x_1, \ldots, x_{n_D}]^T\in \R^{n_D\times d}$ and $\mathbf{Y}_D = [y_1, \ldots, y_{n_D}]^T \in \R^{n_D \times |\mathcal{E}|}$.



\section{Proposed Framework to Investigate the Impact of Unlabeled Data}\label{sec: SS_Framework}

To investigate the impact of incorporating unlabeled samples on event identification performance, and to ensure a fair comparison among various inductive (i.e., self-training) and transductive semi-supervised approaches (i.e., TSVM, LS), we utilize the k-fold cross-validation technique. First, we shuffle $n_{\text{D}}$ samples in $\mathbf{D}$ and partition the data into $n_K$ equally sized folds.  We use $n_K-1$ folds as a training set, denoted as $\mathbf{D}^{(k)}_{T} = \{(x_i, y_i)\}_{i \in \mathcal{I}^{(k)}_{T}}$ with $n_T = \lfloor (n_K - 1) n_D / n_K \rfloor $ samples, and reserve the remaining fold as a validation set, denoted as $\mathbf{D}^{(k)}_{V} = \{(x_i, y_i)\}_{i \in \mathcal{I}^{(k)}_{V}}$ with $n_{V} = n_D - n_T$ samples, and $k = 1, \dots, n_K$. Here, $\mathcal{I}^{(k)}_T$, and  $\mathcal{I}^{(k)}_{V}$ represents a subset of samples in the training set, and the validation set of the $k\Th$ fold, respectively, and $\mathcal{I}^{(k)}_T \cup \mathcal{I}^{(k)}_{V} = \mathcal{I}_D$. We repeat this process $K$ times, with each fold serving as the validation set once.



To further investigate  how the distribution of labeled and unlabeled samples affects the performance of various semi-supervised algorithms, we shuffle the samples in the training set for $n_Q$ times and split it into a subset of $n_L$ labeled samples, denoted as $\mathbf{D}_{L}^{(k,q)} = \{(x_i, y_i)\}_{i \in \mathcal{I}_L^{(k,q)}}$ and a subset of $n_U$ unlabeled samples by ignoring their ground truth labels, denoted as $\mathbf{D}_{U}^{(k,q)} = \{(x_i, \cdot)\}_{i \in \mathcal{I}_U^{(k,q)}}$ where $\mathcal{I}_L^{(k,q)} \cup \mathcal{I}_U^{(k,q)} = \mathcal{I}^{(k)}_T$, and $q = 1, \ldots, n_Q$. 
To ensure the inclusion of samples from every class within the labeled subset, we verify the condition
$B_{\text{min}} \le {n^{c}_L}/{n_L} \le B_{\text{max}}$
where $n^{c}_L$ is the number of samples corresponding to class $c$, and $B_{\text{min}}, B_{\text{max}}$ are the specified balance range.


To illustrate the impact of increasing the number of unlabeled samples, we propose the following procedure. Given the number of samples that we want to add at each step, denoted as \( \Delta_U \), we randomly select \( n_U^{(s)} = s \Delta_U\) from the pool of \( n_U \) samples 
where $ s = 0, \ldots, n_S $, and \( n_S = \lfloor n_U / \Delta_U \rfloor + 1 \) represents the number of steps. To further investigate the impact of the initial distribution of the labeled samples along with the unlabeled samples, the random selection of the \( n_U^{(s)} \) samples at each step $1 \le s \le n_S - 1$, is performed \( n_R \) times.

Concatenating the labeled training samples, $\mathbf{D}_{L}^{(k, q)}$, in the $k$-th fold and $q$-th split, with a subset of $n_U^{(s)}$ unlabeled samples in the $s$-th step and $r$-th random selection ($r \leq n_R$), denoted as $\mathbf{D}_{U}^{(k, q, s, r)} = \{(x_i, \cdot)\}_{i \in \mathcal{I}_{U}^{(k, q, s, r)}}$, where $\mathcal{I}_{U}^{(k, q, s, r)} \subseteq \mathcal{I}_{U}^{(k, q)}$, we obtain a training dataset with mixed labeled and unlabeled samples, denoted as $\mathbf{D}_{M}^{(k, q, s, r)} =\{(x_i, y_i)\}_{i \in \mathcal{I}_L^{(k, q)}} \cup  \{(x_i, \cdot )\}_{i \in \mathcal{I}_{U}^{(k, q, s, r)}}$. 
To account for the semi-supervised learning assumptions, we sort the $n^{(s)}_U$ unlabeled samples in the $\mathcal{I}_{U}^{(k, q, s, r)}$ based on their proximity to the nearest labeled sample.
To improve clarity, for the given $k$, $q$, and $r$, we will modify the superscripts of the training (labeled and unlabeled) and validation samples throughout the remainder of this paper, i.e., $\mathbf{D}_{L}$, $\mathbf{D}^{(s)}_{U}$, $\mathbf{D}^{(s)}_{M}$, and $\mathbf{D}_{V}$ represent the subsets of $n_L$ labeled, $n_U^{(s)}$ unlabeled, $n_M^{(s)} = n_L + n_U^{(s)}$ mixed, and $n_V$ validation samples, respectively. A visual representation of the outlined approach is depicted in Fig. \ref{fig: Framework_orig}.

We can alternatively represent the labeled and unlabeled training samples in matrix format as described below.
We define the matrix of event features with labeled samples as $\mathbf{X}_L = [\dots, x_i, \dots]^T$ and the corresponding matrix of labels as $\mathbf{Y}_L = [\dots, y_i, \dots]^T$ where $i\in \mathcal{I}_L^{(k,q)}$. Similarly, for the subset of unlabeled samples, we define $\mathbf{X}_{U} = [\dots, x_i, \dots]^T$, $i\in \mathcal{I}^{(k, q, s, r)}_{U}$. For the sake of notation coherency as well as implementation considerations (e.g., learning the classification models), we assign value $-1$ to the unlabeled samples, i.e., $\mathbf{Y}_{U} = [-1, \dots, -1]^T \in \R^{n^{(s)}_U}$. Hence, the mixed labeled and unlabeled training set can be expressed as 
 \begin{equation}
      \mathbf{D}_{M}  =\{\mathbf{X}_{M}, \mathbf{Y}_{M}\} 
\end{equation}
where
\begin{equation}
    \begin{split}
         \mathbf{X}_{M} &= [{\mathbf{X}_L}^T, {\mathbf{X}_{U}}^T]^T, \\
          \mathbf{Y}_{M} &= [{\mathbf{Y}_L}^T, {\mathbf{Y}_{U}}^T]^T.
    \end{split}
\end{equation}
Similarly, the validation $\mathbf{D}_{V}$
in the $k\Th$ fold can be represented in the matrix format as $\mathbf{D}_{V} = \{\mathbf{X}_{V}, \mathbf{Y}_{V}\}$ 
 where $\mathbf{X}_{V} = [\dots, x_i, \dots]^T$ and $\mathbf{Y}_{V} = [\dots, y_i, \dots]^T$, and $i\in \mathcal{I}^{(k)}_{V}$.

\section{Semi-supervised Event Identification: \\  Model Learning and Validation }\label{sec: Models}



Our procedure to test semi-supervised methods consists of three steps: (i) pseudo-labeling of unlabeled samples in the training set with mixed labeled and unlabeled samples, $\mathbf{D}^{(s)}_{M}$, (ii) training a classifier using the combined labeled and pseudo-labeled samples, and (iii) evaluating the classifier's performance on the validation set, $\mathbf{D}_{V}$. 

\begin{figure}[b!]
    \centering
    \includegraphics[trim={10 0 0 0 },clip, scale = 0.4]{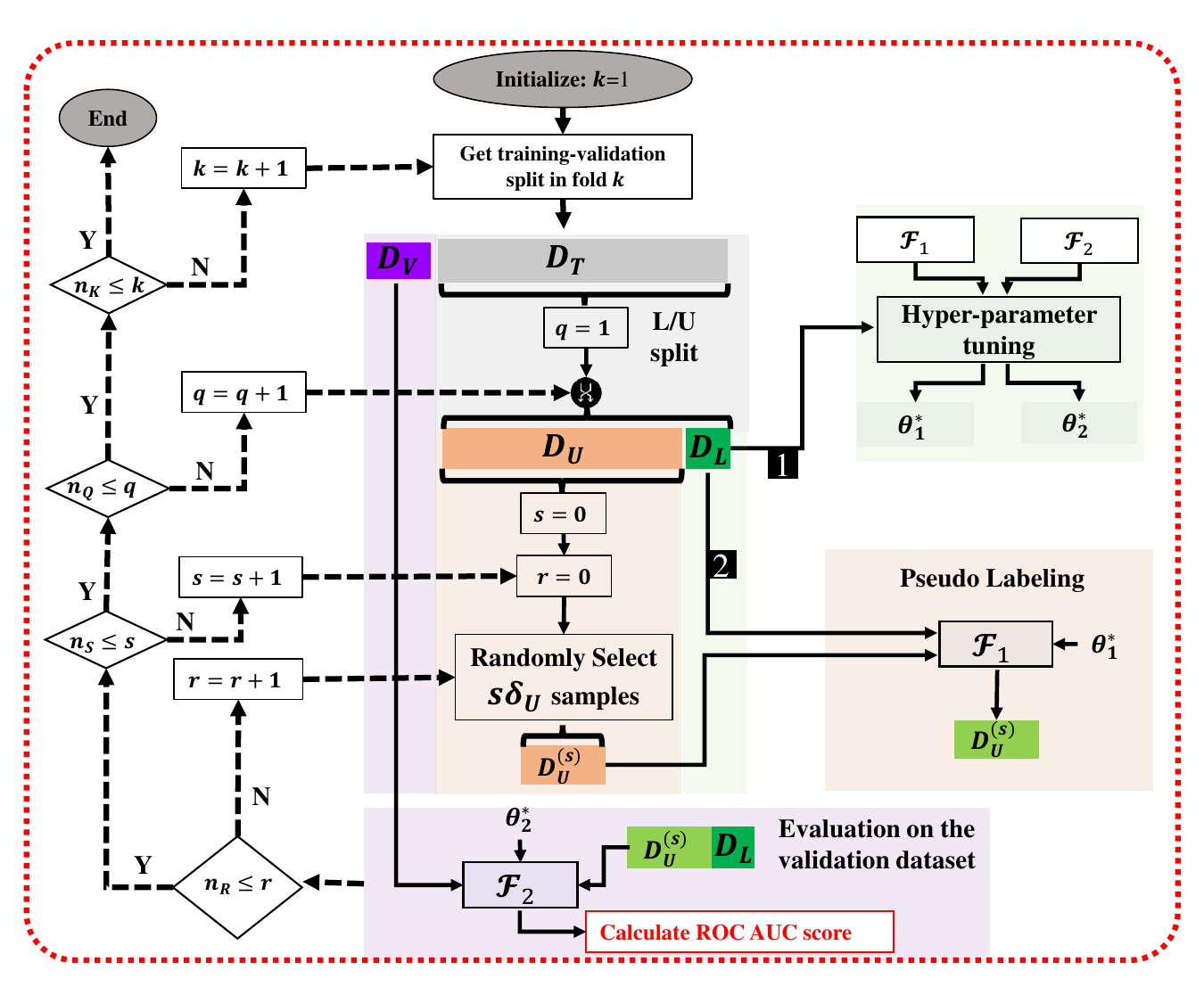}
    \caption{Overview of the proposed semi-supervised pipeline.}
    \label{fig: Framework_orig}
\end{figure}

The overview of the proposed approach is shown in Fig.~\ref{fig: Framework_orig}. Given semi-supervised model $\mathcal{F}_1$ and a classifier $\mathcal{F}_{2}$, we start with the labeled samples within the $k\Th$ fold and the $q\Th$ split of the training set. Using these labeled samples, we perform grid search \cite{gridsearch} to obtain hyperparameters for the models $\mathcal{F}_1$ and $\mathcal{F}_2$, denoted as   $\theta^*_1$ and $\theta^*_{2}$. (Note that these hyperparameters will differ based on $k$ and $q$.) Subsequently, we use the matrix of event features and the corresponding matrix of labels in the $\mathbf{D}^{(s)}_{M}$ to assign pseudo-labels on the unlabeled samples using  $\mathcal{F}_1$. Utilizing the obtained labeled and pseudo-labeled samples, $\widehat{\mathbf{D}}^{(s)}_{M}$, we then use model  $\mathcal{F}_{2} \in \{\text{SVMR, SVML, GB, DT, $K$NN}\}$ to assign labels to the events in the validation dataset $\mathbf{D}_{V}$. 
In the subsequent subsections, we will describe which models we use as $\mathcal{F}_1$ in this procedure.

\subsubsection{Self-training}
Self-training has proven to be effective in leveraging unlabeled data to improve supervised classifiers \cite{self_training_debiased, self1, self2, self3, self4, self5}.
Self-training works by assigning pseudo-labels to unlabeled samples based on the model's predictions and then training the model iteratively with these pseudo-labeled samples. More specifically, for any given base classifier, we learn a model $\mathcal{F}_{1} \in \{\text{SVMR, SVML, GB, DT, $K$NN}\}$ from the labeled samples in the $\mathbf{D}^{(s)}_M$. Then using the learned model, we predict the labels for each $n_U^{(s)}$ unlabeled samples to obtain the augmented labeled and pseudo-labeled samples, denoted as $\widehat{\mathbf{D}}^{(s)}_M$.
Algorithm \ref{alg: self} outlines the steps involved in this procedure. Note that the parameter $\delta_U$ in this algorithm specifies the number of unlabeled samples (among the $n_U^{(s)}$ samples) that will be assigned pseudo-labels in each iteration.

\begin{algorithm}
\caption{Self-Training (for a given  $k, q, s, $ and $r$).}
\begin{algorithmic}[1]
    \State{\textbf{Input: $\mathbf{D}_{M}^{(s)}$}}
    \State{\textbf{Output: $\widehat{\mathbf{D}}^{(s)}_M$}}
    \vspace{0.1cm}
    \State{\textbf{Initialize: } $[\text{f}: \text{t}] = [1: \delta_U]$ \Comment{\textit{\scriptsize \textbf{from sample f to sample t}}}}
    \State{$\quad \qquad \qquad \tilde{\mathbf{X}}_L \leftarrow \mathbf{X}_L, \tilde{\mathbf{Y}}_L \leftarrow \mathbf{Y}_L, \tilde{\mathbf{X}}_U \leftarrow \mathbf{X}_U[\text{f}:\text{t}]$}
    \vspace{0.1cm}
    \While{t $ \le n_U^{(s)}$} 
    \State{$\mathcal{F}_{1}:\tilde{\mathbf{Y}}_{L} \leftarrow \tilde{\mathbf{X}}_{L}$ \Comment{\textit{\scriptsize \textbf{Learning the model}}}}
    \State{$\widehat{\mathbf{Y}}_{U} = \mathcal{F}_{1}(\tilde{\mathbf{X}}_{U})$ \Comment{\textit{\scriptsize \textbf{pseudo-labeling }}}}
    \vspace{0.1cm}
    \State{$\tilde{\mathbf{X}}_L \leftarrow [\tilde{\mathbf{X}}_L^T, \tilde{\mathbf{X}}_{U}^T]^T, \quad \tilde{\mathbf{Y}}_L \leftarrow [ \tilde{\mathbf{Y}}_L^T, \widehat{\mathbf{Y}}_{U}^T]^T$ \Comment{\textit{\scriptsize \textbf{Augmentation}}}}
    \vspace{0.1cm}
    \State{$f \leftarrow f + \delta_U, \quad t \leftarrow t + \delta_U$}
    \State{\textbf{if} t > $n_U^{(s)}$: 
    \State{$\quad$ t $ = n_U^{(s)}$} }
    \State{$\tilde{\mathbf{X}}_U \leftarrow \mathbf{X}_U[\text{f}:\text{t}]$}
    \EndWhile
    \State{$\widehat{\mathbf{Y}}_{M} \leftarrow \tilde{\mathbf{Y}}_L$}
    \State{\textbf{Return:  }  $\widehat{\mathbf{D}}^{(s)}_M  =\{\mathbf{X}_{M}, \widehat{\mathbf{Y}}_{M}\}$}
\end{algorithmic}\label{alg: self}
\end{algorithm}

\subsubsection{Transductive Support Vector Machine (TSVM)}
  The TSVM approach is a modification of the SVM formulation that addresses the challenge of limited labeled data in classification tasks \cite{QSTSVM_method, TSVMorig, surveySSL}. The TSVM optimization problem is given by 
\begin{subequations}\label{eq:tsvm}
\begin{equation}
\min_{\mathbf{w}, b, \boldsymbol{\eta}, \boldsymbol{\zeta}, \mathbf{z}} \quad C \left[ \sum_{i\in \mathcal{I}_L} \eta_i + \sum_{j\in \mathcal{I}_U} \min(\zeta_j, z_j) \right] + \|\mathbf{w}\|^2
\end{equation}
subject to:
\begin{align}
& y_i (\mathbf{w}^T x_i - b) + \eta_i \geq 1, \quad \eta_i \geq 0, \quad i \in \mathcal{I}_L \\
& \mathbf{w}^T x_i - b + \zeta_j \geq 1, \quad \zeta_j \geq 0, \quad j \in \mathcal{I}_U \\
& -(\mathbf{w}^T x_i - b) + z_j \geq 1, \quad z_j \geq 0, \quad j \in \mathcal{I}_U
\end{align}
\end{subequations}
 where $\mathbf{w} \in
\mathbb{R}^d$ and $b \in \mathbb{R}$ represent the direction of the decision boundary and the bias (or intercept) term, respectively. It introduces two constraints (i.e., (5c), and (5d)) for each sample in the training dataset calculating the misclassification error as if the sample belongs to one class or the other. The objective function aims to find $\mathbf{w}$ and $b$ that, while maximizing the margin and reducing the misclassification error of labeled samples (i.e., $\boldsymbol{\eta}$), minimize the minimum of these misclassification errors (i.e., $\boldsymbol{\zeta}$ and $\mathbf{z}$). 
 This enables the TSVM to utilize both labeled and unlabeled samples for constructing a precise classification model. Subsequently, it assigns pseudo-labels to the unlabeled samples. 
For brevity, we refer readers to \cite{QSTSVM_method, TSVMorig} for more comprehensive details.

\subsubsection{Label Spreading (LS)}
    
Label spreading (LS) falls within the category of graph-based semi-supervised (GSSL) models \cite{GSSL_survey}. It involves constructing a graph and inferring labels for unlabeled samples where nodes represent samples and weighted edges reflect similarities. 
Consider a graph $G_{M}=(\mathcal{V}_{M}, \mathcal{W}_{M})$ which is constructed over the combined labeled and unlabeled training set.
Each sample, $x_i,  \forall i \in \mathcal{I}_L \cup \mathcal{I}_U$, 
can be represented as a node in a graph. 
For the resulting graph, we define the edge weights matrix as $\mathcal{W}_M \in \R^{n^{(s)}_M\times n^{(s)}_M}$. Defining $D_{ij} = \lVert x_i-x_j \rVert^2$, the $i\Th$ row and $j\Th$ column of $\mathcal{W}_M$, 
denoted as $w_{ij}$, can be obtained as $w_{ij} = \exp({- D_{ij}/2\sigma^2})$ if $i\neq j$, and $w_{ii}=0$. For such a measure of edge weight, proximal pairs of samples will have larger weights. 
Building on the classical intuition that proximal samples tend to have the same labels, the LS approach enables propagation of labels from the labeled to unlabeled samples through weighted edges where the weights carry the notion of similarity. 
In Algorithm~\ref{alg: lp}, we detail the steps of the LS approach based on \cite{LSzhou}. The update rule is captured in line 7 in Algorithm~\ref{alg: lp} wherein the labels for both the labeled and unlabeled samples are updated; in particular, for the labeled samples, such an update includes information from the neighbors (first term) while preserving the initial label (second term). The parameter $\alpha$ determines the weighting between neighbor-derived information and the sample's original label information.


\begin{algorithm}
\caption{Label spreading (for a given  $k, q, s, $ and $r$).}
\begin{algorithmic}[1]
    \State{\textbf{Input: $G = (\mathcal{V}, \mathcal{W})\leftarrow \mathbf{D}^{(s)}_{M}  =\{\mathbf{X}_{M}, \mathbf{Y}_{M}\}$}}
    \State{\textbf{Output: $\widehat{\mathbf{D}}^{(s)}_{M}$}}
    \vspace{0.1cm}
    \State{\textbf{Compute: $\mathcal{D}_{ii} = \sum_{j} w_{ij} , \quad \forall i \in \mathcal{I}_L \cup \mathcal{I}_{U}$}}
    \vspace{0.1cm}
    \State{\textbf{Compute: $\mathbf{Z} = \mathcal{D}^{-1/2}\mathcal{W}_{M}\mathcal{D}^{-1/2}$}}
    \vspace{0.1cm}
    \State{\textbf{Initialize: $\begin{bmatrix} \mathbf{Y}_L|_{t=0} \\ \mathbf{Y}_{U}|_{t=0} \end{bmatrix} \leftarrow \begin{bmatrix} \mathbf{Y}_L \\ \mathbf{Y}_{U} \end{bmatrix}$}}
    \vspace{0.1cm}
    \While{$\begin{bmatrix} \mathbf{Y}_L|_{t} \\ \mathbf{Y}_{U}|_{t} \end{bmatrix}$ converges} \Comment{\textit{\scriptsize Based on some threshold}}
    \State{$\begin{bmatrix} \mathbf{Y}_L|_{t+1} \\ \mathbf{Y}_{U}|_{t+1} \end{bmatrix} \leftarrow \alpha \mathbf{Z} \begin{bmatrix} \mathbf{Y}_L|_{t} \\ \mathbf{Y}_{U}|_{t} \end{bmatrix} + (1-\alpha) \begin{bmatrix} \mathbf{Y}_L|_{t=0} \\ \mathbf{Y}_{U}|_{t=0}\end{bmatrix}$}
    \vspace{0.1cm}
     \State{$t \leftarrow t+1$}
    \EndWhile  \label{LP loop}
    \vspace{0.1cm}
    \State{$\widehat{\mathbf{Y}}_{M} \leftarrow \begin{bmatrix} \mathbf{Y}_L|_{t} \\ \mathbf{Y}_{U}|_{t} \end{bmatrix}$}
    \State{\textbf{Return:}  $\widehat{\mathbf{D}}^{(s)}_{M}  =\{\mathbf{X}_{M}, \widehat{\mathbf{Y}}_{M}\}$}
\end{algorithmic}\label{alg: lp}
\end{algorithm}




\section{Simulation Results}\label{sec: sim_result}
In order to investigate the performance of various semi-supervised learning algorithms, we first generate eventful synthetic PMU data, following the procedure described in Section \ref{sec: II_data_generation}. Our simulations were carried out on the South-Carolina 500-Bus System \cite{TX500_1, TX500_2}. We allow the system to operate normally for $t_f = 1$ second and then we immediately apply a disturbance.
We then run the simulation for an additional $t_s=10$ seconds, and 
record the resulting eventful measurements at the PMU sampling  rate of $30$ samples/sec. The $t_{\text{clr}}$ for the BF events is 5 cycles ($\approx 0.083$ seconds). 
We assume that 95 buses (which are chosen randomly) of the Carolina 500-bus system are equipped with PMU devices and extract features for each such bus from the $V_m$, $V_a$, and $F$ channels. We thus collect $N=300$ samples after the start of an event for each channel. We use the modal analysis methodology as outlined in our recent prior work \cite{NT_TPS} to extract features using modal analysis.
In total, we simulated $1827$ events including $500$ LL, $500$ GL, $500$ LT, and $327$ BF events. 


To quantitatively evaluate and compare the performance of different semi-supervised learning algorithms across various scenarios, we employ the area under curve (AUC) of the receiver operator characteristic (ROC). This metric enables the characterization of the accuracy of classification for different discrimination thresholds.
The ROC AUC value, which ranges from 0 to 1, provides an estimate of the classifier's ability to classify events. A value of AUC closer to 1 indicates a better classification performance.
For a specified set of parameters $k$, $q$, $s$, and $r$, we evaluate the performance of a given classifier $\mathcal{F}_2$ by assessing its ROC-AUC score in predicting event classes within the hold-out fold. This evaluation is based on the model learned from the augmented labeled and pseudo-labeled samples, which are obtained using the pseudo-labeling model $\mathcal{F}_1$.

Given that the aim of this study is to provide insight into the robustness of various semi-supervised models, we compare them by evaluating the average, $5\Th$ percentile, and $95\Th$ percentile of the AUC scores based on the accuracy of the assigned pseudo-labels on the unlabeled samples and assess the impact of incorporating the assigned pseudo-labels on the accuracy of a generalizable model in predicting the labels of validation samples.
We use the $5\Th$ percentile of the AUC scores as our primary target performance metric for robustness, as it provides a (nearly) worst-case metric across different selections of the initial labeled and unlabeld samples. That is, if a method yields a high $5\Th$ percentile performance, then it is likely to lead to accurate results, even if the initial set of labeled and unlabeled samples are unfavorable.
Within this setting, to ensure a fair comparison among various inductive and transductive semi-supervised approaches, we consider two distinct approaches:
\begin{itemize}
    \item \textbf{Approach 1 (Inductive semi-supervised setting): } \\ $\mathcal{F}_{1} \in \{\text{SVMR, SVML, GB, DT, $K$NN}\}$ represents the base classifier utilized in self-training for pseudo-labeling, and the same type of classifier will be used as $\mathcal{F}_{2}$.
    \item \textbf{Approach 2 (Transductive semi-supervised setting): } \\ $\mathcal{F}_{1} \in \{\text{TSVM, LS}\}$ represents a semi-supervised method used for pseudo-labeling, and  $\mathcal{F}_{2} \in \{\text{SVMR}, \text{SVML}, \text{GB},$\\$ \text{$K$NN}\}$.
\end{itemize}

\begin{figure*}[h]
    \centering
    \begin{subfigure}{0.49\textwidth}
        \includegraphics[trim={30 20 30 20 },clip, width=\linewidth]{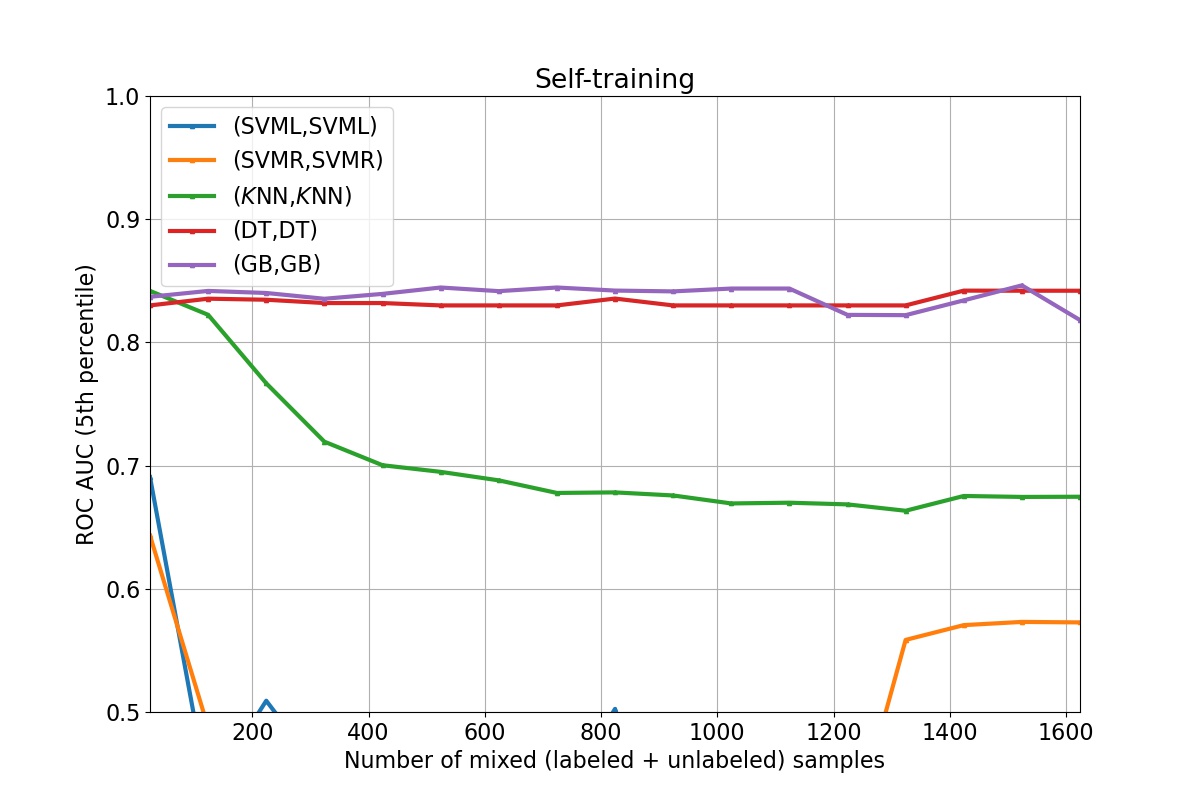}
        \caption{}
        \label{fig:subfig1}
    \end{subfigure}
    \begin{subfigure}{0.49\textwidth}
        \includegraphics[trim={30 20 30 20 },clip, width=\linewidth]{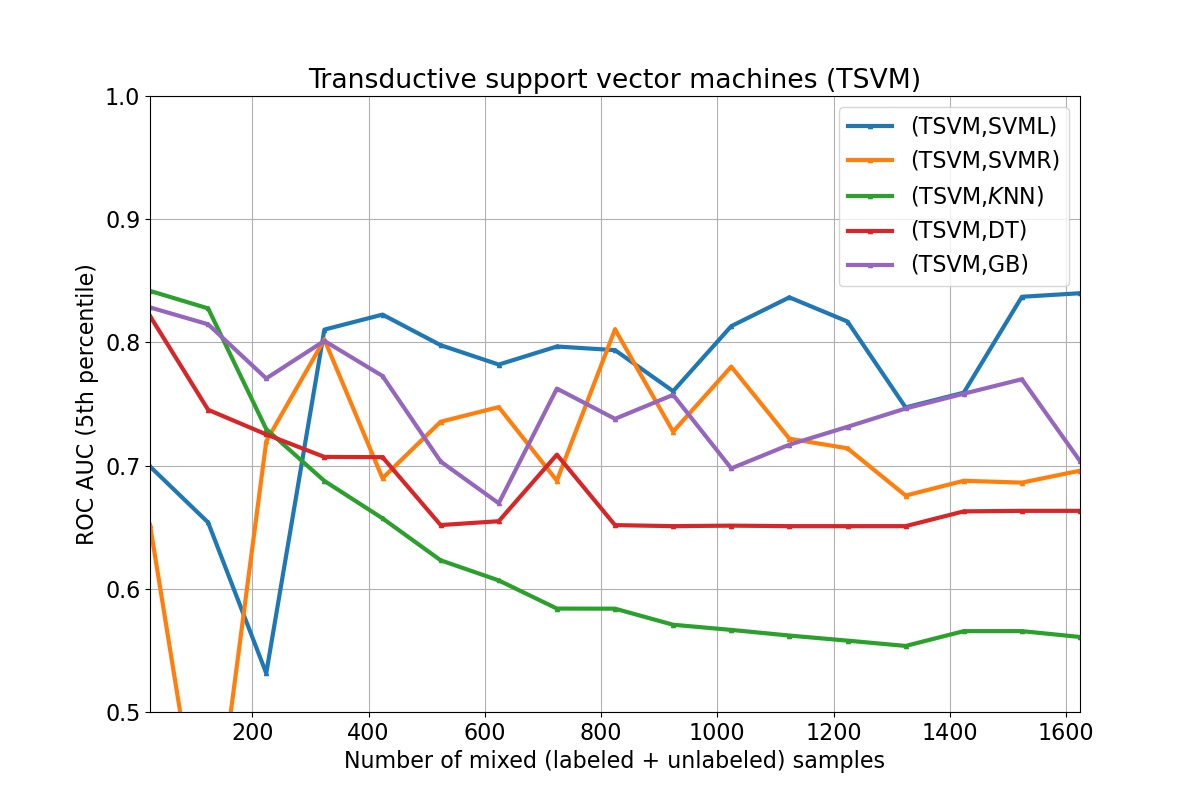}
        \caption{}
        \label{fig:subfig2}
    \end{subfigure}
    \begin{subfigure}{0.49\textwidth}
        \includegraphics[trim={30 20 30 20 },clip, width=\linewidth]{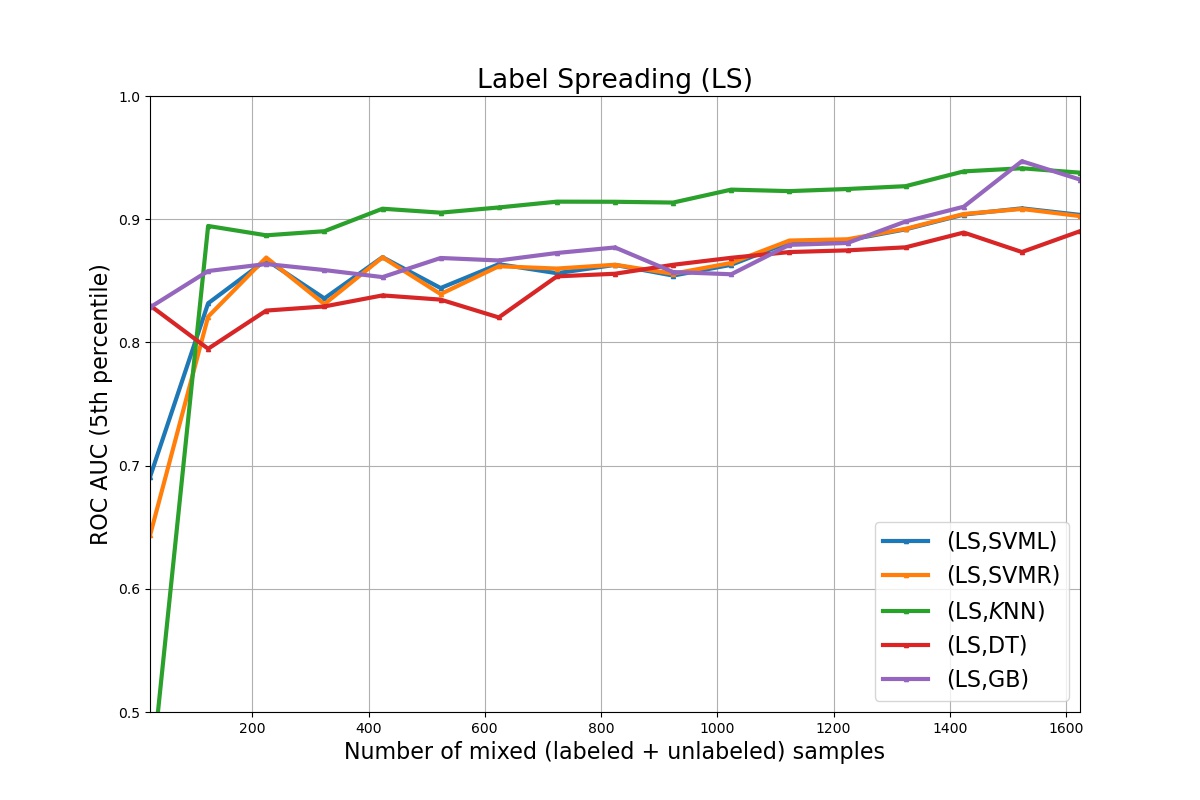}
        \caption{}
        \label{fig:subfig3}
    \end{subfigure}
    \begin{subfigure}{0.49\textwidth}
        \includegraphics[trim={30 32 30 20},width=\linewidth]{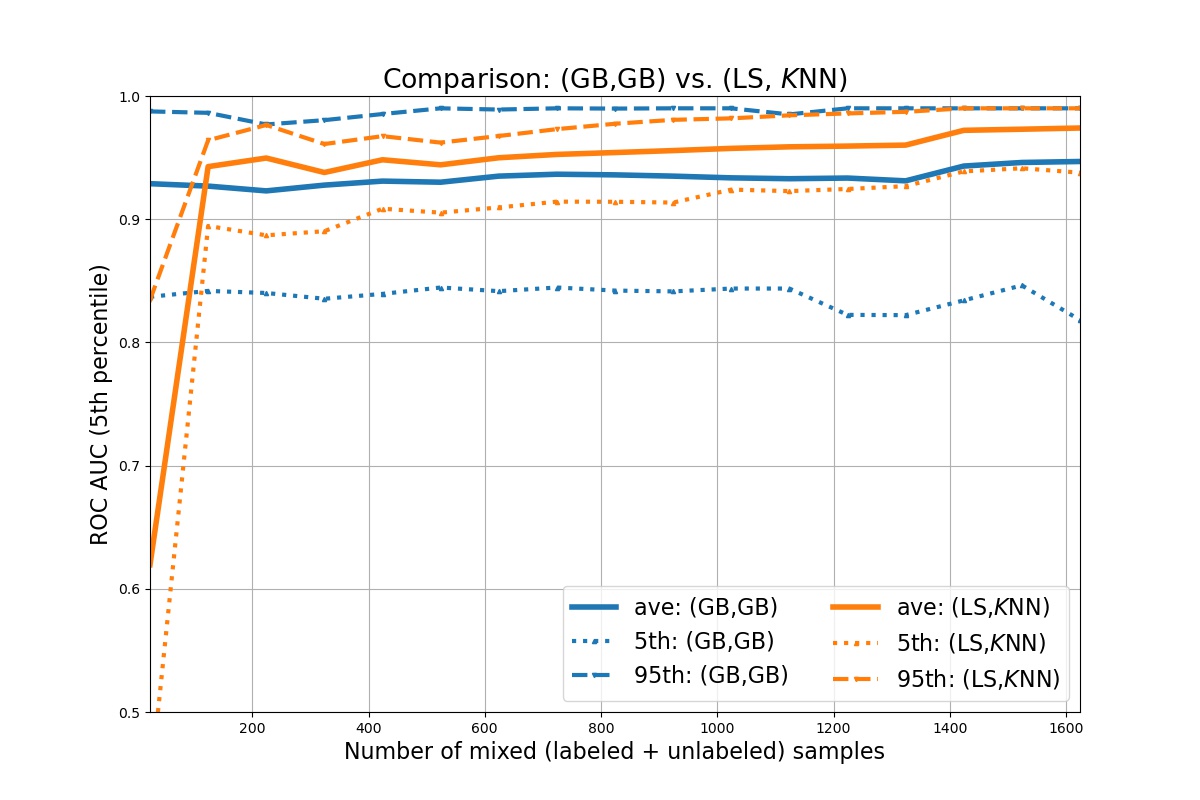}
        \caption{}
        \label{fig:subfig4}
    \end{subfigure}
    \caption{The $(\mathcal{F}{1}, \mathcal{F}{2})$ pairs denote the selection of pseudo-labeling and validation classifiers.The $5\Th$ percentile of AUC scores for different classifiers using pseudo-labels obtained from: (a) Self-training method with various base classifiers, (b) TSVM, and (c) LS. 
(d) Comparison between (GB, GB) and (LS, $K$NN) in terms of average, $5\Th$, and $95\Th$ percentile of AUC scores. } 
    \label{fig:main4}
\end{figure*}

 In our evaluation process, we take into account $n_K = 10$ folds and $n_Q = 30$ random splits of the training samples into labeled and unlabeled subsets. 
 As discussed in Sec. \ref{sec: Models}, we perform hyperparameter tuning of the models using the labeled training samples within each fold. 
  Other simulation parameters, and models hyperparameter values are provided in Tables \ref{table:sim_setting} and \ref{table:sim_setting2} respectively. For the LS model, we utilize the same $\gamma$ value as obtained from the hyperparameter tuning of the SVMR model. 
As depicted in Figure \ref{fig:main4}, the comparative performance of diverse classifiers (namely, SVML, SVMR, $K$NN, DT, and GB) is presented across distinct semi-supervised models (self-training, TSVM, and LS). The outcomes of this analysis highlight that the integration of additional unlabeled samples and the utilization of LS for pseudo-labeling surpasses the outcomes achieved by the self-training and TSVM approaches.  Moreover, the LS algorithm consistently enhances the performance of all classifiers more robustly. The following subsections provides further insight on the performance of each semi-supervised model.

\begin{table}[b]
\caption{Parameters used in semi-supervised event identification}\label{table:sim_setting}
\begin{tabular}{|l|l|l|}

\hline
\textbf{Parameter  }                            &\textbf{ Description}                                                                                                             &\textbf{ Value}      \\ \hline
$n_D$                                  & Total no. of samples                                                                                                    & 1827      \\ \hline
$n_K$                                  & No. of folds                                                                                                             & 10         \\ \hline
$n_T$                                  & No. of training samples                                                                                                 & 1644       \\ \hline
$n_{V}$                          & No. of validation samples                                                                                                     & 183         \\ \hline
$n_Q$                                  & \begin{tabular}[c]{@{}l@{}}No. of random splits of training samples \\ into labeled and unlabeled  samples\end{tabular} & 20         \\ \hline
$(B_{\text{min}}, B_{\text{max}})$     & \begin{tabular}[c]{@{}l@{}}Class balance range in the\\ labeled samples\end{tabular}                                    & (0.2, 0.8) \\ \hline
$n_L$                                  & No. of labeled samples                                                                                                  &  24        \\ \hline
$n_U$                                  & No. of Unlabeled samples                                                                                                & 1620       \\ \hline
$\delta_U$                             & \begin{tabular}[c]{@{}l@{}}No. of unlabeled samples in each step\end{tabular}                                        & 100        \\ \hline
$n_S$                                  & Total No. of steps                                                                                                      & 18          \\ \hline
$n_R$ & \begin{tabular}[c]{@{}l@{}} No. of random selection\\ of $n^{(s)}_U$ samples at each step\end{tabular}      & 10  \\ \hline
\end{tabular}
\end{table}

\vspace{0pt}

\begin{table}[]
\caption{Values used for hyperparameter tuning of the models in semi-supervised event identification.}\label{table:sim_setting2}
\begin{tabular}{|c|c|c|}
\hline
\textbf{Model}        & \textbf{Hyperparameter}             & \textbf{Values}               \\ \hline
KNN                   & No. of neighbors in $K$NN           & 2, 4, 6, 8, 10                \\ \hline
SVML                  & Regularization parameter            & logspace($10^{-3}, 10^2, 10$)$^*$ \\ \hline
\multirow{2}{*}{SVMR} & $\gamma$ in RBF kernel               & logspace($10^{-3}, 10^2, 10$) \\ \cline{2-3} 
                      & Regularization parameter            & logspace($10^{-3}, 10^2, 10$) \\ \hline
DT                    & Maximum depth                       & 3, 5, 7                       \\ \hline
\multirow{2}{*}{GB}   & No. of estimators (boosting stages) & 50, 100, 150, 200             \\ \cline{2-3} 
                      & Maximum depth                       & 3, 5, 7                       \\ \hline
\end{tabular}
\footnotesize{$^*$ logspace($a, b, n$) represents $n$ values in a logarithmic grid from $a$ to $b$}
\end{table}
\vspace{0pt}

\subsection{Approach 1 --- Inductive semi-supervised setting }
The simulation results for the $5\Th$ percentile of the AUC scores of the SVML, SVMR, $K$NN, DT, and GB classifiers in predicting the labels of validation samples are shown in Fig.~\ref{fig:subfig1}.  
It is clear that using a limited number of labeled samples, results in poor performance for the self-training method when utilizing SVMR, SMVL, and $K$NN base classifiers. Moreover, the utilization of GB and DT as base classifiers does not necessarily lead to an improvement in event identification accuracy. This primarily arises from the disparity between the pseudo-labels and the initial subset of labeled samples. Training with biased and unreliable pseudo-labels can result in the accumulation of errors. In essence, this pseudo-label bias exacerbates particularly for classes that exhibit poorer behavior, such as when the distribution of labeled samples does not accurately represent the overall distribution of both labeled and unlabeled samples, and is further amplified as self-training continues.

Another noteworthy observation is that self-training employing SVML or SVMR as the classifiers exhibits a high sensitivity to the distribution of both labeled and unlabeled samples. Due to the constraint of having a limited number of labeled samples, these techniques struggle to generate dependable pseudo-label assignments. On the other hand, although self-training with $K$NN as the base classifier performs better than SVML and SVMR cases, its performance deteriorates as we increase the number of the unlabeled samples. 
For the self-training with DT and GB base classifiers, it is evident that, although they exhibit more robust performance compared to other types of base classifiers, increasing the number of unlabeled samples does not enhance their performance.

\subsection{Approach 2 --- Transductive semi-supervised setting }
The simulation results for the second approach in which TSVM is employed as the semi-supervised method for pseudo-labeling are illustrated in Fig. \ref{fig:subfig2}.
The weak performance of TSVM could be attributed to the specific characteristics of the dataset and the method's sensitivity to the distribution of labeled and unlabeled samples. If the distribution of these samples is unbalanced or exhibits complex patterns, the TSVM might struggle to accurately capture this distribution. As a result, it could assign inaccurate pseudo-labels. Furthermore, it becomes evident that the integration of pseudo-labels acquired through the TSVM algorithm, although yielding an overall performance advantage for SVML and SVMR when compared to the same models utilizing pseudo-labels from the self-training algorithm involving SVMR and SVML, still exhibits substantial sensitivity. This sensitivity is particularly apparent when assessing the $5\%$ AUC scores, highlighting that the accuracy of assigned pseudo-labels remains highly contingent on the initial distribution of labeled and unlabeled samples. This phenomenon is also observable in the diminishing performance of the $K$NN, GB, and DT classifiers, which, surprisingly, deteriorates to a level worse than their utilization as base classifiers within the self-training framework.

On the contrary, as shown in Fig. \ref{fig:subfig3},  the results demonstrate that utilizing the augmented labeled and pseudo-labeled samples obtained from LS can significantly enhance the performance of event identification, as compared to the self-training and TSVM approaches.  Furthermore, the performance of the event identification task improves with a higher number of unlabeled samples, which is particularly significant since labeled eventful PMU data is often scarce in practice. 
The principal advantage of the LS method, when compared to self-training and TSVM, primarily arises from its ability to leverage information from both labeled and unlabeled samples, as well as their inherent similarities, during the assignment of pseudo-labels. 
For some classifiers (specifically GB and DT), we find that LS improves the $5\Th$ percentile line with more unlabeled samples, even though the average performance stays roughly unchanged. On the other hand, for the $K$NN classifier (as shown in Fig. 3d), the average, $5\Th$, and $95\Th$ percentile lines all improve with more unlabeled samples. Indeed, LS with $K$NN seems to be the best overall classifier.

\section{Conclusion}\label{sec: s6_Conclusion}


Given the practical scenario where a relatively small number of events are labeled in comparison to the total event count, we have introduced a semi-supervised event identification framework to explore the potential benefits of incorporating unlabeled samples in enhancing event identification performance. This framework comprises three core steps: (i) assigning pseudo-labels to unlabeled samples within the training set, which encompasses a mixture of labeled and unlabeled samples, (ii) training a classifier using the augmented set of labeled and pseudo-labeled samples, and (iii) evaluating the classifier's efficacy on the holdout fold. This proposed pipeline is deployed to scrutinize the effectiveness of three classical semi-supervised methods: self-training, TSVM, and LS. 
Our simulation results suggests that using a limited number of labeled samples, the self-training and TSVM methods perform poorly and does not necessary improve the accuracy of event identification.
The study underscores the robust performance of GB and DT classifiers, though augmenting unlabeled samples does not enhance their performance. 
 Conversely, using the augmented labeled and pseudo-labeled samples obtained from LS consistently outperform the self-training and TSVM approaches, and can significantly improve event identification performance. The performance also improves with a higher number of unlabeled samples, which is important given the scarcity of labeled eventful PMU data. 
 




\balance
\bibliographystyle{IEEEtran}
\bibliography{References.bib}

\end{document}